\pdfoutput=1
\documentclass[11pt]{article}

\usepackage[preprint]{acl}

\usepackage{times}
\usepackage{latexsym}

\usepackage[T1]{fontenc}
\usepackage[utf8]{inputenc}

\usepackage{microtype}

\usepackage{inconsolata}

\usepackage{graphicx}

\usepackage{hyperref}
\usepackage{url}
\usepackage{booktabs}
\usepackage[flushleft]{threeparttable}
\usepackage{subcaption}
\usepackage{colortbl}
\usepackage{makecell}
\usepackage{multirow} 

\title{Patent-CR: A Dataset for Patent Claim Revision}

\author{
Lekang Jiang$^{\dagger}$,  
Pascal A Scherz$^{\diamond}$,
Stephan Goetz$^{\dagger}$ \\
$^{\dagger}$University of Cambridge, $^{\diamond}$PSPB Patent Law\\ 
\texttt{\{lj408, smg84\}@cam.ac.uk, post@pspb.eu}
}

\begin{document}
\maketitle
\begin{abstract}

This paper presents Patent-CR, the first dataset created for the patent claim revision task in English. It includes both initial patent applications rejected by patent examiners and the final granted versions.
Unlike normal text revision tasks that predominantly focus on enhancing sentence quality, such as grammar correction and coherence improvement, patent claim revision aims at ensuring the claims meet stringent legal criteria. These criteria are beyond novelty and inventiveness, including clarity of scope, technical accuracy, language precision, and legal robustness. 
We assess various large language models (LLMs) through professional human evaluation, including general LLMs with different sizes and architectures, text revision models, and domain-specific models. 
Our results indicate that LLMs often bring ineffective edits that deviate from the target revisions.  
In addition, domain-specific models and the method of fine-tuning show promising results.
Notably, GPT-4 outperforms other tested LLMs, but further revisions are still necessary to reach the examination standard.
Furthermore, we demonstrate the inconsistency between automated and human evaluation results, suggesting that GPT-4-based automated evaluation has the highest correlation with human judgment.
This dataset, along with our preliminary empirical research, offers invaluable insights for further exploration in patent claim revision.\footnote{\url{https://github.com/scylj1/Patent-CR}}

\end{abstract}

\section{Introduction}

Text revision aims to improve text quality, such as fixing grammar errors \citep{fang2023chatgpt} and enhancing sentence coherence \citep{geva2019discofuse}. Currently, datasets for this task are derived from scientific literature, Wikipedia entries, and news articles \citep{du2022understanding}. In this paper, we broaden the scope of text revision to encompass the domain of patents, characterized by large-scale, complex, and precise textual data. The patent domain presents unique opportunities and challenges for the field of artificial intelligence (AI) and natural language processing (NLP)~\citep{jiang2025natural}. 

\begin{figure*}[!t]
    \centering   
    \includegraphics[width=\textwidth]{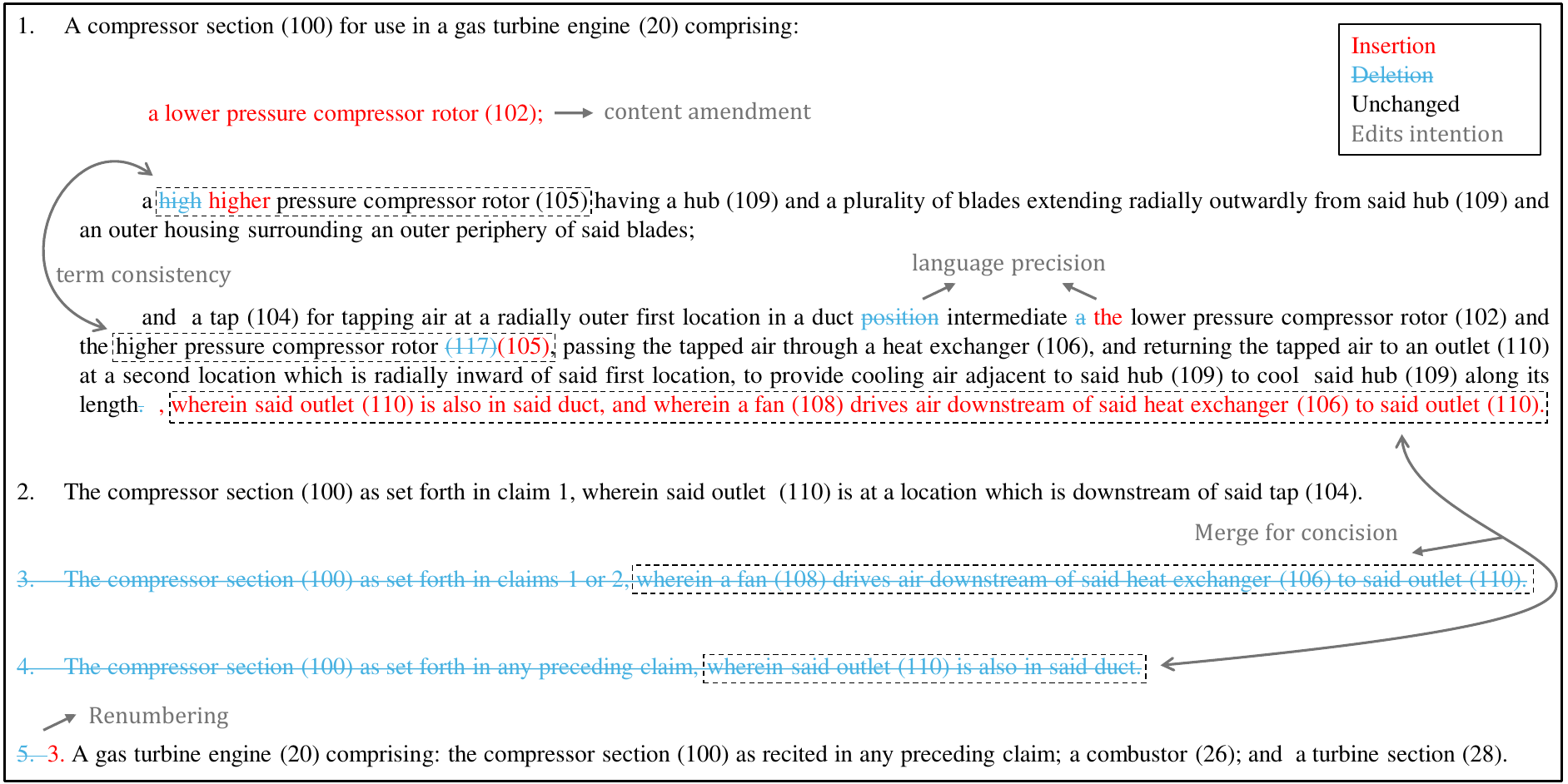}   
    \caption{An example of claim revision for patent EP3181869}
    \label{fig:claim}
\end{figure*}

Patent claims are critical in a patent application document. As the legal centerpiece, they define the technical scope of the invention and ensure the patent can withstand legal scrutiny. We introduce the relevant background information of patents in Appendix~\ref{background}. Drafting and revising patent applications are both time-intensive and financially burdensome \citep{CisloAndThomas2023}. Research showed that large language models (LLMs) have the potential to generate high-quality patent claims but current performance is not yet satisfactory \citep{jiang-etal-2025-large, epd2025}. To facilitate the automation of patent writing, we propose a new task, namely patent claim revision, aiming at improving the quality of patent claims to pass the legal scrutiny of patent offices. 

Figure~\ref{fig:claim} illustrates an example of patent claim revision. By checking the dataset and consulting patent professionals, we have identified five different types of modifications between the draft and final versions. 
\textbf{(1)	Content amendment:} Essential information missing in the draft is included to highlight novel features of the invention, and unnecessary or redundant information is removed, including overlaps with the prior art.
\textbf{(2)	Term consistency:} Technical terms are ensured to be consistent throughout the document.
\textbf{(3)	Language precision:} Grammatical errors are corrected, and word choice is refined for greater precision.
\textbf{(4)	Concision:} Some claims are merged with others for concision. 
\textbf{(5)	Renumbering:} The consolidation of claims necessitates adjustments in their numbering.

Our main contributions are detailed as follows: 

\noindent 1. We introduce a novel task namely patent claim revision and present the first dataset for research and evaluation, comprising 22,606 pairs of application and published claims originating from the same patent.  

\noindent 2. We conduct an empirical study with professional human assessments to evaluate different LLMs on this task. Our findings reveal that most LLMs tend to simply input claims, leading to deviations from intended revisions. Additionally, domain-specific models and fine-tuning demonstrate promising performance. Moreover, despite the best performance of GPT-4 among all tested LLMs, its outputs remain substantially below the desired standard, highlighting the inherent complexity and challenge of accurately revising patent claims.

\noindent 3. We assess the correlation between automated and human evaluations, revealing that GPT-4-based evaluations correlate most closely with human judgments. Developing new automated evaluation metrics that better align with human assessments can be a promising future research direction.

\section{Related Work}

\begin{table*}[!ht]
\centering
\footnotesize
%\resizebox{\textwidth}{!}{
\begin{tabular}{l|c|c|c}
\toprule
\textbf{Dataset} & \textbf{Size} & \textbf{Domain} & \textbf{Granularity} \\
\midrule
ArgRewrite \citep{zhang-etal-2017-corpus} & 180 & Academic & Sentence \\
\citet{anthonio-etal-2020-wikihowtoimprove} & 2.7M & Wikipedia & Sentence \\
NewsEdits \citep{spangher2021newsedits} & 4.6M & News & Sentence \\
ITERATER \citep{du2022understanding} & 31K & Scientific articles, Wikipedia, and news & Sentence \& Paragraph  \\
CASIMIR \citep{jourdan2024casimir} & 3.7M & Scientific articles & Sentence \\
\rowcolor{gray!30}
Patent-CR (Ours) & 22.6K & Patent claims & Paragraph  \\
\bottomrule
\end{tabular}
%}
\caption{Comparison with previous text revision datasets. }
\label{table:datasets}
\end{table*}

\subsection{Text Editing}

Text editing entails the modification of input texts to serve various objectives. This field has historically concentrated on tasks such as correcting grammatical errors \citep{fang2023chatgpt}, paraphrasing \citep{chowdhury2022novelty}, simplifying text \citep{vstajner2022sentence}, and transferring writing styles \citep{reif2022recipe}. Previously, researchers fine-tuned LLMs using datasets comprising original and modified texts without specific instruction-tuning \citep{faltings2021text, kim2022improving}. Inspired by groundbreaking efforts in fine-tuning LLMs based on human-written instructions \citep{ouyang2022training, longpre2023flan}, researchers have begun to explore instruction-tuned models also for text revision. For instance, \citet{schick2023peer} fine-tuned T5-based LLMs for text editing by incorporating human-provided text-editing plans. Furthermore, \citet{raheja2023coedit} explored the capacity of instruction-tuned LLMs to handle complex and multi-part instructions for text editing. More recently, \citet{jourdan2024casimir} introduced a novel dataset specifically designed for revisions of scientific articles. We compare our dataset with previous text revision datasets in Table \ref{table:datasets}. Our dataset broadens the scope of text revision to the patent domain. 

\subsection{Patent Writing}

The adoption of LLMs in generating patent text primarily aims to enhance the efficiency and efficacy of drafting patent applications. Despite the potential capabilities of LLMs, current research in this area remains limited and largely unsatisfactory \citep{jiang2025natural}. An early study by \citet{lee2020patentgenerate} served as a preliminary exploration into generating patent claims with the fine-tuning of GPT-2 \citep{radford2019language}. The authors demonstrated that minimal training steps were adequate for the model to generate patent-like texts, but they did not evaluate the quality of the generated text. Subsequent research by \citet{lee2020controlling} expanded on this aspect by training GPT-2 to convert one element of a patent application into another, for example, creating abstracts from titles and claims from abstracts. As the abstract is typically rather generic and imprecise, the latter may not be a well-conditioned task. Hence, \citet{jiang-etal-2025-large} proposed the description-based claim generation task and evaluated the performance of the current LLMs on this domain-specific task. We extend the task to claim revision to explore whether LLMs can further improve the quality of claims. 
Moreover, \citet{christofidellis2022pgt} introduced a prompt-based generative transformer (PGT) for patent-related tasks, which used GPT-2 as a foundational model and employed multi-task learning (MTL) \citep{maurer2016benefit} to train on various tasks, including part-of-patent generation, text infilling, and evaluating patent coherence. Additionally, \citet{aubakirova2023patfig} presented the first large-scale patent figure-caption dataset, designed for patent figure caption generation.

\section{Dataset}

\subsection{Data Collection}

\begin{figure*}[!t]
    \centering   
    \includegraphics[width=\textwidth]{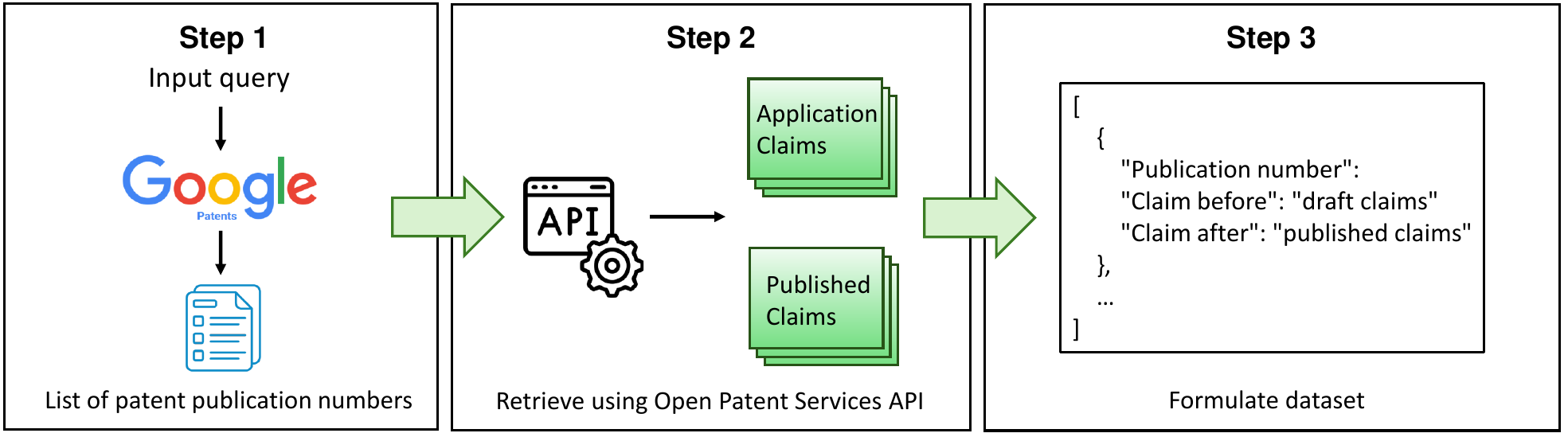}   
    \caption{Steps to create the dataset}
    \label{fig:process}
\end{figure*}

Figure~\ref{fig:process} demonstrates the three steps through which we collected and created the dataset.  

\noindent \textbf{Step 1: }
Firstly, we searched for published and granted patents using advanced search options on Google Patents.\footnote{\url{https://patents.google.com/}} We set \textit{Language} to English, \textit{Patent Office} to European Patent Office, \textit{Status} to Grant, and \textit{Type} to Patent. We downloaded a list of patent publication numbers that contain European patents published from January 2024 to June 2024. A patent publication number is a unique identifier assigned to a patent application when published, which was used for claim text retrieval in further steps. We have chosen the latest sets of patents to minimize the possibility that the LLMs have been trained on those texts. 

\noindent \textbf{Step 2: }
The European Patent Office provides the Open Patent Services (OPS) for public access to their data. We retrieve the application and published versions of claims corresponding to specific patent publication numbers through the OPS API. A patent has different versions published by EPO, where A1 or A2 is the patent application and B1 is the granted patent.
We primarily used the A1 version, and we used A2 if A1 is not available. We eliminated the patents that do not have either A1 or A2. We opted for the B1 version as the revised claims and discarded those without the B1 version. 

\noindent \textbf{Step 3: }
In the final compilation of our dataset, we formulate data into an easy-readable format and manually check in detail to ensure the dataset's quality. 

\begin{table}[t!]
\centering
\footnotesize
\begin{tabular}{l|c|c}
\toprule
 & \textbf{Original} & \textbf{Revised} \\
\midrule

\multicolumn{2}{l}{\textbf{Statistics}} \\

\# Documents & 22,606 & 22,606  \\
\# Claims & 13.85 & 10.66  \\
\# Tokens & 1,391 & 1,285 \\
Claim length & 101 & 121 \\
Structure complexity & 1.05 & 1.44  \\
Readability ($\downarrow$) & 30.18 & 37.24 \\
\midrule

\multicolumn{2}{l}{\textbf{Changes}} \\
Total edits & \multicolumn{2}{c}{619}  \\
Addition  & \multicolumn{2}{c}{238}  \\
Deletion & \multicolumn{2}{c}{353}  \\
Replacement  & \multicolumn{2}{c}{28} \\

\bottomrule
\end{tabular}
\caption{Data statistics of our Patent-CR dataset. The methods to calculate these statistics are introduced in Appendix~\ref{statscalculation}. A smaller value of readability score indicates higher readability. The number of edits is calculated at the word level, representing the average number of word changes per patent document. }
\label{table:dataset_statistics}
\end{table}

\subsection{Dataset Information}

This dataset comprises 22,606 pairs of initial and published claims. 
Table \ref{table:dataset_statistics} shows the data statistics of the Patent-CR dataset. On average, draft claims consist of 13.85 claims and 1,391 tokens, while published claims feature 10.66 claims and 1,285 tokens. This reduction in both claims and tokens in published versions underscores a trend towards enhanced conciseness and/or the integration of dependent claims into independent ones to establish novelty or inventiveness over prior art with additional features. Notably, structure complexity increases from 1.05 to 1.44 and the readability score rises from 30.18 to 37.24, where a higher score indicates reduced readability. These findings underscore a pivotal aspect of patent claim revision: the revised claims become more complex and less readable. This contrasts with normal text revision tasks, which generally aim to enhance readability. More dataset statistics are included in Appendix~\ref{morestats}.

\section{Experiments}

We selected the patent claims in June 2024 as the test set and the remaining ones as the training set for fine-tuning or few-shot prompting. We introduce the experimental details in Appendix \ref{experimentsdetail}, including the prompts and environmental settings.

\subsection{Models}

To make a comprehensive evaluation, we select various models for experiments. A detailed introduction of these models is reported in Appendix~\ref{modeldetail}. 

We use the original claims without any editing as a baseline similar to previous text revision works \citep{raheja2023coedit, jourdan2024casimir}, namely the \textbf{Copy} baseline. This baseline performs well when there are extensive overlaps between source inputs and target outputs. We can also evaluate other models' performance by comparing the results with this simple baseline. We select the state-of-the-art \textbf{CoEdIT-XL}~\citep{raheja2023coedit} as the representative of text revision models. As the patent is a form of legal document, legal-specific LLMs may be useful in this task. Hence, we evaluate \textbf{SaulLM-7B}, a model designed for the legal domain \citep{colombo2024saullm}. For general open-source LLMs, we opt for two types of models with different structures.  We include \textbf{Mixtral-8$\times$7B}~\citep{jiang2024mixtral} based on Sparse Mixture of Experts (SMoE) and the recent Llama-3.1 series~\citep{dubey2024llama}. We evaluate both the \textbf{Llama-3.1-8B} and \textbf{Llama-3.1-70B} versions to explore the size effect. In addition, we use LoRA \citep{hu2021lora} to fine-tune our own model, \textbf{Llama-3.1-8B-FT} and \textbf{SaulLM-7B-FT}, to investigate the effectiveness of fine-tuning. Fine-tuning details are introduced in Appendix \ref{experimentsdetail}. We also test the powerful GPT series for comparison, including \textbf{GPT-3.5} and state-of-the-art \textbf{GPT-4}~\citep{achiam2023gpt}.

\subsection{Evaluation}

\noindent \textbf{Human Evaluation }
To enhance the precision of evaluation for this new task, we incorporate evaluations by patent professionals, adhering strictly to established examination criteria. Recent research~\citep{jiang-etal-2025-large} suggests five criteria for assessing the quality of patent claims, which align perfectly with our patent claim revision objectives introduced in the Introduction section. 

\textbf{(1) Completeness of Essential Features} (score 1 – 10): The extent to which the generated claims encapsulated all critical aspects of the invention.
It corresponds to our revision goal of content amendment, ensuring that essential information missing in the draft is included or unnecessary details are removed.
\textbf{(2) Conceptual Clarity} (score 1 – 10): The clarity and unambiguity of the language used in the claims.
It reflects our revision goal of enhancing language precision by correcting grammatical errors and refining word choice.
\textbf{(3) Consistency in Terminology} (score 1 – 10): The uniformity in the use of terms throughout the claims.
It matches our goal of maintaining consistent technical terminology across the document.
\textbf{(4) Technical Correctness of Feature Linkages} (score 1 – 10): The accuracy with which the features were interconnected and related.
It relates to our aims of improving concision (by merging some claims) and renumbering (to adjust claim numbering as necessary).
\textbf{(5) Overall Quality} (score 1 – 10): An aggregate measure combining all the above criteria.
$Quality = (Completeness*4 + Clarity*2 + Consistency*2 + Linkage*3) \div 11$

Given the high cost and labor intensity of involving patent experts in evaluating a large number of claim sets, we conducted human evaluations on a select set of 60 examples (6 examples for each model's outputs). Patent professionals compared the referenced claims with those generated by LLMs, rating each on the aforementioned criteria on a scale from 1 to 10, where a higher score indicates better performance.

\noindent \textbf{Automated Evaluation }
We use the standard metrics for text revision, including \textbf{SARI} \citep{xu2016optimizing}, \textbf{BLEU} \citep{papineni2002bleu}, \textbf{ROUGE-L (R-L)} \citep{lin2004rouge}, and \textbf{BERTScore} \citep{zhang2019bertscore}. Appendix~\ref{standardeval} introduces details of these metrics. Previous research has also applied the Exact Match (EM) metric for evaluating text revisions \citep{raheja2023coedit, jourdan2024casimir}, which quantifies the proportion of candidate texts that exactly match reference texts. However, EM is not applicable in our case because the process of revising patent claims does not adhere to strict one-to-one correspondence. The quantity of claims may vary between the original and revised texts.

Moreover, studies have shown that LLM-based evaluators can achieve better human alignment \citep{liu-etal-2023-g}. Thus, we use GPT-4 with Chain-of-Thought (CoT) \citep{wei2022chain} prompting to evaluate generated patent claims, namely \textbf{G-Eval}. GPT-4 is given the human evaluation criteria, claims being evaluated, and reference claims. We ask GPT-4 to evaluate the given claims step-by-step and assign a score for each criterion. The detailed settings and prompts are introduced in Appendix~\ref{gevaldetails}. We do not use GPT-4 to evaluate the outputs generated by itself because they may be biased.

\noindent \textbf{Statistics }
To investigate the characteristics of different models' outputs, we also count some statistical information for comparison, including the averaged \textbf{number of tokens, number of claims, claim length, structure complexity, and readability}. The methods to calculate these statistics are introduced in Appendix~\ref{statscalculation}.

\section{Results and Discussion}

\begin{table*}[!t]
\centering
\footnotesize
\resizebox{\textwidth}{!}{
\begin{tabular}{l|ccccc|cccccc}
\toprule
\multirow{2.7}{*}{\textbf{Model}} & \multicolumn{5}{c|}{\textbf{Automated Evaluation}} & \multicolumn{5}{c}{\textbf{Human Evaluation}} \\
\cmidrule(lr){2-6} \cmidrule(lr){7-11}
 & SARI & BLEU & R-L & BERTScore & G-Eval & Completeness & Clarity & Consistency & Linkage & Quality \\
\midrule
Copy & \underline{59.9} & \underline{0.63} & \underline{0.68} & \underline{0.92} & 80.7 & 5.67 & 5.50 & 5.83 & 5.33 & 5.58 \\
CoEdIT-XL & 34.6 & 0.59 & 0.64 & 0.91 & 76.8 & 5.17 & 4.82 & 5.16 & 4.67 & 4.97 \\
SaulLM-7B & 42.6 & 0.51 & 0.61 & 0.91 & \textbf{\underline{81.8}} & 5.50 & 5.50 & 5.83 & 5.50 & 5.56 \\
SaulLM-7B-FT & 55.1 & \textbf{\underline{0.63}} &  \textbf{0.67} & \textbf{\underline{0.92}} & 80.7 & 6.33 & \underline{\textbf{6.50}} & \underline{\textbf{6.67}} & 6.17 & 6.38 \\
Mixtral-8x7B & 33.2 & 0.27 & 0.47 & 0.88 & 81.7 & 5.33 & 5.17 & 5.67 & 5.17 & 5.32 \\
Llama-3.1-8B & 38.4 & 0.48 & 0.54 & 0.90 & 79.4 & 5.33 & 5.33 & 5.17 & 5.17 & 5.26 \\
Llama-3.1-8B-FT & \textbf{55.5} & 0.62 & 0.66 & \textbf{\underline{0.92}} & 80.3 & 5.83 & 6.17 & 6.33 & 6.00 & 6.03 \\
Llama-3.1-70B & 38.7 & 0.49 & 0.56 & 0.90 & 78.1 & 5.83 & 5.67 & 5.83 & 5.17 & 5.62 \\
GPT-3.5 & 38.2 & 0.49 & 0.60 & 0.90 & 76.9 & 5.67 & 5.67 & 5.83 & 5.33 & 5.60 \\
GPT-4 & 33.7 & 0.45 & 0.55 & 0.89 & - & \underline{\textbf{6.67}} & 6.17 & 6.17 & \underline{\textbf{6.33}} & \underline{\textbf{6.40}} \\
\bottomrule
\end{tabular}
}
\caption{Evaluation results of different models. The best result of each metric is \underline{underlined} and the best result among models for each column is in \textbf{bold}. We do not use G-Eval to evaluate outputs of GPT-4 as it may be biased. The values of G-Eval are the overall quality and we report the full results of G-Eval in Table~\ref{table:g_evaluation}. In automated evaluation results, the copy baseline shows strong performance and the fine-tuned model outperforms other LLMs. GPT-4 shows the best performance on human evaluation metrics. }
\label{result}

\end{table*}

\begin{table*}[!t]
\centering
\footnotesize
\resizebox{.7\textwidth}{!}{
\begin{tabular}{l|c|c|c|c|c}
\toprule
\textbf{Claim Texts}           & \textbf{\# Tokens} & \textbf{\# Claims} & \textbf{Length} & \textbf{Complexity} & \textbf{Readability $\downarrow$} \\ \midrule
Reference  & 958 & 9.76 & 98 & 1.21 & 34.40 \\           
Copy & 1,124 & 13.56 & 83 & 0.94 & 26.77 \\ 
CoEdIT-XL  & 1,039 & 13.56 & 77 & 1.37 & 24.61 \\
SaulLM-7B  & 756 & \textbf{11.48} & 66 & \textbf{1.13} & 25.84 \\
SaulLM-7B-FT  & 1032 & 12.34 & 84 &  0.90 & \textbf{38.32} \\
Mixtral-8$\times$7B & 1,492 & 11.73 & 127 & 2.10 & 22.72 \\
Llama-3.1-8B & 1,220 & 13.08 & \textbf{93} & 1.66 & 23.36 \\   
Llama-3.1-8B-FT & 1,106 & 12.98 & 85 & 0.92 & 28.70 \\ 
Llama-3.1-70B & 1,085 & 13.56 & 80 & 1.55 & 21.63 \\
GPT-3.5    & 831 & 13.94 & 60 & 0.67 & 22.78 \\
GPT-4      & \textbf{891} & 14.01 & 63 & 0.77 & 23.31 \\ 
\bottomrule
\end{tabular}
}
\caption{Statistics of gold referenced claims, original copy of claims, and model output claims. The results are the averaged numbers of all evaluated texts. The value in each column closest to the value of referenced claims is marked in \textbf{bold}. A smaller value of readability score indicates higher readability.}
\label{stat}

\end{table*}

We report the empirical results of automated and human evaluations in Table~\ref{result}. We primarily focus on human assessment outcomes, but automated metrics also provide valuable insights at times. To further elucidate the aims of the patent claim revision and examine the behaviors of various models, we list other statistical information in Table~\ref{stat}. Based on the synthesis of results from two tables, we provide insightful observations and a comprehensive result analysis.

\subsection{Challenges for LLMs on Patent Claim Revision}
The copy of original claims reaches high scores in automated evaluation metrics with a SARI of 59.9, BLEU of 0.63, R-L of 0.68, and BERTScore of 0.92, as shown in Table~\ref{result}. The result implies a significant overlap between source and target texts, suggesting that the original claims are largely accurate and require minimal modifications. However, patent claims must be exceptionally clear and precise without ignoring any tiny mistakes. Thus, further revisions are essential and demand meticulous attention, improving the complexity of the task. 

We compare the statistical difference between the original copy and reference claims to analyze the goals of patent claim revision and its difference from normal text revision tasks.  As shown in Table~\ref{stat}, there is a reduction in the average number of tokens from 1,124 to 958 and a decrease in the number of claims from 13.56 to 9.76. Nonetheless, the average claim length increases from 83 to 98, indicating denser and more succinct target contents. Conversely, the reference text exhibits increased structural complexity (1.21 compared to 0.94) and reduced readability (with readability scores of 34.40 compared to 26.77, where a higher score denotes lower readability). This increase in complexity is inherently different from normal text revision tasks, where target texts are usually more readable. The differences can be attributed to the specialty of patent claims. Normal text revision often aims to enhance clarity, coherence, readability, etc., typically through simplification of structure and shortening of clauses. By contrast, patent claim revisions prioritize unambiguity, technical precision, and legal robustness to meet patent office criteria. Patent claims also use standardized legal and technical language, where every term and phrase has a potential legal implication, necessitating a focus on accuracy and consistency. This specialized focus renders patent claim revision substantially more challenging than conventional text editing tasks. Furthermore, the addition of features from dependent claims to the independent ones to differentiate the invention from the prior art increases sentence length and may in some cases add further clauses, e.g., relative clauses.

With respect to G-Eval and human evaluation metrics, the copy baseline also outperforms some LLMs. Most small-sized LLMs struggle to make substantial improvements to original patent claims. In human evaluations, the quality of some revised claims generated by LLMs does not surpass the baseline quality score of 5.58 of original claims, such as Llama-3.1-8B (5.26) and CoEDIT-XL (4.97). This result suggests a tendency of such models to deviate significantly from the gold standard in revising the original claims, leading to ineffective edits. A possible reason is that despite the valuable content of the patent literature, these models are not pre-trained on large-scale patent data, resulting in the models' inability to capture special linguistic features of patent texts. Furthermore, GPT-4 shows the highest human evaluation quality of 6.40, but it is insufficient to pass the patent examination. 

\textbf{Takeaways } 
Three challenges complicate patent claim revision. (1) The original claims are substantially accurate, necessitating only minimal revisions. Using LLMs to refine these claims for perfection is difficult. (2) Patent texts exhibit unique linguistic characteristics, posing challenges for general LLMs. (3) Unlike normal text revision objectives, the primary goal of patent claim revision is to align with specific patent criteria.

\subsection{Ineffectiveness of General Text Revision Models}
Although CoEdIT is the state-of-the-art model for normal text revision, its application to patent claim revision yields unsatisfactory outcomes. 
As reported in Table~\ref{result}, CoEdIT reaches the lowest human evaluation quality score of 4.97 and G-Eval score of 76.8 among all tested models. In addition, the scores on all metrics are below the original copy's performance, demonstrating the model's inability to make meaningful edits. This highlights significant limitations in applying general text revision techniques to the specialized field of patent language.

From Table~\ref{stat}, we observe that CoEdIT tends to decrease token count (from 1,124 to 1,039) and claim length (from 83 to 77) while increasing readability. These amendments are expected because CoEdIT was originally designed to improve text quality like readability. As illustrated in the above section, the purpose of claim revision is inherently different from normal text revision and the task is more difficult. Therefore, it is understandable that CoEdIT underperforms when not trained on patent-specific texts.

Another limitation in applying current text revision models such as CoEdIT to patent claim revision is their short context length. These models often support (short-) sentence-level edits. For example, CoEdIT has an input length of 256 tokens, which is significantly less than the approximate 1,000-token average length of a patent claim set. This limitation necessitates processing each claim individually without altering the total number of claims, whereas optimal revision would consider the patent claims collectively, aiming for conciseness and precision through content integration. Patent claim revision, therefore, is more accurately described as a paragraph-level editing task, requiring simultaneous processing of multiple sentences. 

\textbf{Takeaways } Current general text revision models are not suitable for patent claim revision. Training such models on patent texts and increasing the context length may increase the performance.

\subsection{Results of Law-Specific LLM}
SaulLM-7B, a model specifically tuned for legal text, shows promise by achieving a quality score of 5.56, outperforming similar-sized general LLMs, such as Llama-3.1-8B with a score of 5.26. 
Moreover, Table~\ref{stat} shows that claims generated by SaulLM have the closest number of claims and structure complexity to the gold claims. This model benefits from training on a blend of patent data and extensive legal texts, which appears to enhance its ability to adhere to standard patent language requirements, such as consistent terminology usage. The overall quality of SaulLM-7B is comparable to that of much larger general models like Llama-3.1-70B and GPT-3.5, underscoring the potential benefits of domain-specific training. 

\textbf{Takeaways } SaulLM-7B outperforms similar-sized general LLMs, suggesting that law-specific or patent-specific LLMs may achieve better performance. Research could focus on expanding the size of these models and training them with more diverse legal and patent datasets, including international patent laws and multilingual patent databases. In addition, investigating adaptive learning techniques that allow LLMs to continuously update their training as they are exposed to new patent filings and legal precedents could help maintain their relevance and accuracy over time.

\begin{table*}[!ht]
\centering
\footnotesize
\resizebox{.95\textwidth}{!}{
\begin{tabular}{l|cc|cc|cc|cc|cc}
\toprule
\multirow{2.7}{*}{\textbf{Metric}}    & \multicolumn{2}{c|}{\textbf{Completeness}} & \multicolumn{2}{c|}{\textbf{Clarity}} & \multicolumn{2}{c|}{\textbf{Consistency}} & \multicolumn{2}{c|}{\textbf{Linkage}} & \multicolumn{2}{c}{\textbf{Quality}} \\ \cmidrule{2-11}
          & $\rho$      & $\tau$      & $\rho$    & $\tau$   & $\rho$      & $\tau$     & $\rho$    & $\tau$   & $\rho$    & $\tau$   \\ \midrule
SARI      & 0.153         & 0.107            & 0.350       & 0.265         & 0.127         & 0.095           & 0.258       & 0.187         & 0.246       & 0.162         \\
BLEU      & 0.523         & 0.411            & \textbf{0.627}       & \textbf{0.513}         & 0.220         & 0.170           & \textbf{0.574}       & \textbf{0.440}         & 0.577       & 0.426         \\
R-L   & 0.505         & 0.406            & 0.589       & 0.477         & 0.169         & 0.134           & 0.488       & 0.379         & 0.520       & 0.392         \\
BERTScore & 0.298         & 0.218            & 0.499       & 0.385         & \textbf{0.246}         & \textbf{0.188}           & 0.425       & 0.320         & 0.408       & 0.278         \\
G-Eval & \textbf{0.624}         & \textbf{0.527}            & 0.576       & 0.507         & 0.172         & 0.132           & 0.530       & 0.435         & \textbf{0.600}       & \textbf{0.444}         \\ \bottomrule
\end{tabular}
}
\caption{Spearman ($\rho$) and Kendall-Tau ($\tau$) correlation of automated evaluation with human evaluation results. The highest number in each column is in \textbf{bold}. G-Eval is most related to human evaluations of claims' overall quality.}
\label{table:correlation}
\end{table*}

\subsection{Advantages of Fine-tuning}
Table~\ref{result} illustrates that fine-tuned Llama-3.1-8B achieves the highest SARI (55.5), while fine-tuned SaulLM-7B reaches the best BLEU (0.63) and R-L (0.67) among all LLMs in automated evaluations. In human evaluation, the fine-tuned models outperform their corresponding base models and also the copy baseline in all aspects, demonstrating the effectiveness of fine-tuning. This finding aligns with previous research on patent claim generation \citep{jiang-etal-2025-large}. Particularly, SaulLM-7B-FT achieves almost the same overall quality score (6.38) as GPT-4 (6.40), and it even outperforms GPT-4 in clarity and consistency. This finding suggests that in-domain training would bring significant advantages to patent text generation, a specific type of text featuring high precision. 

\textbf{Takeaways } Fine-tuning leads to improvements across all evaluation metrics compared to the original model. Researchers with sufficient computational resources could investigate the efficacy of full-parameter fine-tuning or extend these methods to larger LLMs.

\subsection{Outstanding Performance of GPT-4}

In human evaluations, GPT-4 stands out by generating the most qualified claims among all tested models, with an overall quality improvement from 5.58 to 6.40. Although the outputs are short, they include more essential invention features, increasing the completeness. Notably, it is the only model that shows a marked improvement in the feature linkage, rising from 5.67 to 6.67. GPT-4 effectively reorganizes different embodiments of the invention in a logical manner, enhancing the connections between features, whereas other models can not. Nonetheless, the quality score of 6.40 is not enough to pass rigorous patent examination. Therefore, despite the advancements, the claims produced by GPT-4 still require further refinement to meet the stringent standards of patent scrutiny. 

Table~\ref{stat} indicates that the claims generated by GPT-4 have less structure complexity and better readability compared to the copy baseline. This pattern indicates that GPT-4 also tends to simplify the original texts, which is the possible reason that GPT-4 achieves low scores on lexical evaluation.  

\textbf{Takeaways } Although GPT-4 outperforms other tested LLMs, the generated claims still need further revision. Moreover,  GPT-4 is the only model that can reorganize different invention features logically to improve the correctness of feature linkage. Future research may focus on developing or integrating models that specialize in causal and logical reasoning. This would help LLMs understand and apply the underlying logical structures that are crucial for accurately linking and grouping patent claim features.

\subsection{Inconsistency between Automated and Human Evaluations}

We can observe from Table~\ref{result} that the automated evaluation metrics may not be well-suited for this patent task. This is testified by the strong performance of the simple copy baseline and the poor performance of GPT-4 on automated evaluation. To further investigate this issue, we present the Spearman ($\rho$) and Kendall-Tau ($\tau$) correlation between automated evaluation and human evaluation results in Table~\ref{table:correlation}. We use the \textit{scipy} Python library to calculate the correlation scores. 

SARI has the least correlation with all human evaluation criteria. SARI evaluates the presence or absence of certain words and phrases (additions, deletions, and copies). However, LLMs may significantly reconstruct the original sentences, such as modifying sentence structures, changing word orders, and replacing words with synonyms. If those modifications deviate from the target lexical revision, the SARI scores are low, but in fact, the revised claims may have better quality. In addition, BERTScore is also ineffective in patent claim revision because semantic information is almost unchanged in this task. Claims generated by each LLM have a similar BERTScore, making it difficult to differentiate the actual claims' quality. BLEU and R-L show a relatively higher correlation with human evaluations except for term consistency. This suggests that the lexical overlap with gold claims can to some extent reflect the quality of generated claims. 

G-Eval shows the most robust performance on overall quality, especially in feature completeness with Spearman and Kendall-Tau correlation of 0.624 and 0.527 respectively. This suggests that GPT-4 can capture and compare invention features from patent claims effectively. However, G-Eval's results on other sub-criteria, particularly terminology consistency, are not outstanding. It is worth noting that none of the automated metrics can achieve over 0.25 correlation with human evaluation in consistency, which can be an interesting research direction for the future. Overall, G-Eval generally outperforms other metrics, with the best Spearman correlation of 0.600 and Kendall-Tau correlation of 0.444 in evaluating the quality. 

\textbf{Takeaways } The inconsistency of the automated metrics with the human gold standard shows the limitations of the metrics. G-Eval is a currently more suitable choice to automatically evaluate patent claims. There is still a need for better automated evaluation methods for patents that have closer alignment to human expert evaluation. A follow-up work investigated a new evaluation method specifically designed for patent claims \citep{jiang2025towards}.

\section{Conclusion}

We introduce the first dataset for English patent claim revision namely Patent-CR, providing valuable resources for research and evaluation in this newly proposed task. Our empirical study of various cutting-edge LLMs and professional human evaluations reveal the inherent challenges of the task. Most small-scale LLMs predominantly simplify inputs, deviating from the target purpose of claim refinement. Law-specific models and the fine-tuning of general LLMs show promising performance. Although GPT-4 outperforms other LLMs on this task, the output claims still need further refinement to meet stringent examination criteria, underscoring the task's complexity. Additionally, we point out the inconsistency between automated and human evaluation results, suggesting that GPT-4-based evaluation has the highest correlation with human assessment.
Consequently, the patent claim revision task presents multiple challenges that necessitate resolution for advancements in this field.

\section*{Limitations}
The dataset is restricted to patents published by the European Patent Office and documented in English. We do not conduct hyper-parameter tuning when doing experiments. 

\section*{Ethics Statement}
Llama-3 is under \textit{META LLAMA 3 COMMUNITY LICENSE AGREEMENT}. GPT-4 is under a commercial license provided by OpenAI, and we access it through its API. Our dataset is collected from the EPO's Open Patent Services (OPS). According to rule 3.1 in \textit{Terms and Conditions for use of the EPO's OPS}, users may use and include these data in their own machine-readable databases, products and services ("products") and may distribute the data as part of these products. This dataset does not include potential personal information and offensive content. The use of existing artifacts is consistent with their intended use. Our proposed dataset is used for patent claim revision and released under \textit{CC-BY-SA-4.0} license.

\bibliography{anthology,custom}
%\bibliography{custom}

\appendix

\section{Patent Background}
\label{background}

Patent documents are distinct from normal texts, posing specific challenges for text revision. Firstly, patent language is highly specialized, incorporating technical terminology, legal phrases, and sometimes novel terms to describe new concepts that may not yet be widely acknowledged. This specialized language presents considerable difficulties for general-purpose large language models (LLMs) trained on standard texts, such as Wikipedia. For example, technical jargon may cause issues for current tokenizers because these terms may not appear during model pre-training. Furthermore, language models might struggle to accurately interpret the context of patents as the important terms can be completely new to the language models or have different meanings compared to everyday texts.  
Secondly, patent texts must be precise to ensure the patent is both defensible and enforceable. A technical term must not be substituted with synonyms unless explicitly indicated as equivalent within the patent document itself. This requirement for precision in patent texts complicates the task of generating patent-specific content.

Patent claims are critical in a patent application document. As the legal centerpiece, they define the technical scope of the invention and ensure the patent can withstand legal scrutiny. The description, on the other hand, is rather the dictionary and explanation for the claim. Claims must be written with precision and clarity, as they define the technical matter that should be protected and should contain the key atomic elements of the gist of the invention, i.e., the features that constitute the inventor's novel and inventive technology. Generally, patent claims are categorized into two types: independent claims and dependent claims. Independent claims describe the essential features of an invention without relying on any other claims. They aim to cover the invention as broadly as possible, encompassing various implementations and variations, while remaining specific enough to distinguish it from prior art. Dependent claims, attached to an independent claim, introduce additional features, i.e., limitations to refine a specific embodiment or variant of the invention.  

Drafting patent claims typically requires the expertise of professional patent agents or lawyers, given its requirement for an in-depth grasp of the invention's technical nuances, as well as familiarity with patent laws and writing conventions. A correct and precise definition of patent claims is the key to securing robust patent protection. However, the processes of drafting and revising patent applications are both time-intensive and financially burdensome, posing significant challenges, particularly for small enterprises aiming to engage with the intellectual property (IP) system. Consequently, a smart digital patent writing assistant could markedly enhance the quality and efficiency of the drafting process. Furthermore, the automation of patent drafting has the potential to foster technological innovation and bolster the technological development of society. To facilitate the automation of patent writing, we propose a new task, namely patent claims revision, aiming at improving the quality of patent claims to pass the legal scrutiny of patent offices. 

\begin{figure*}[!t]
    \centering

    \begin{subfigure}[]{0.48\textwidth}
        \includegraphics[width=\textwidth]{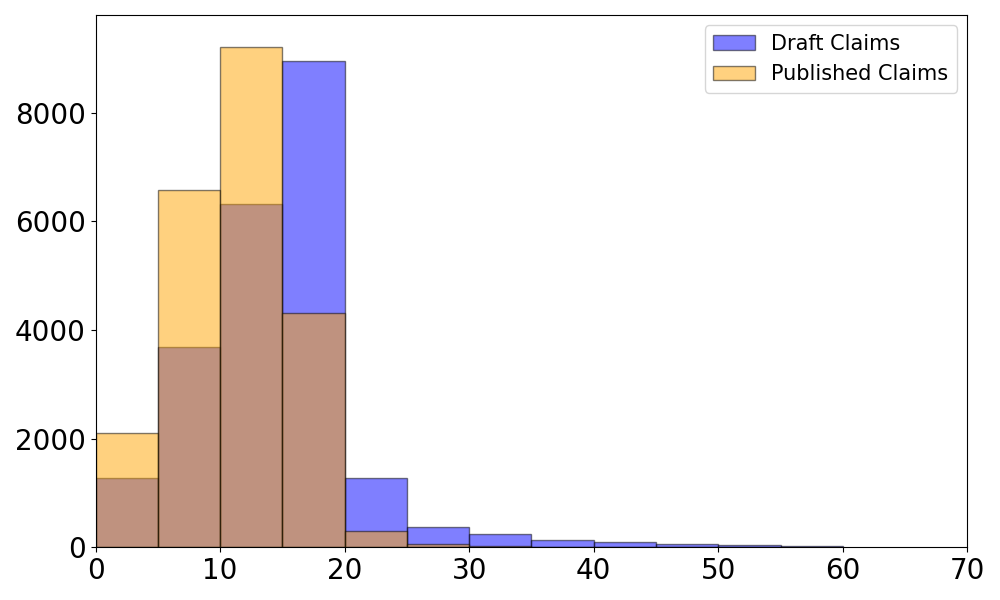}
        \caption{Number of draft claims and published claims}
        \label{fig:1}
    \end{subfigure}
    \hfill   
    \begin{subfigure}[]{0.48\textwidth}
        \includegraphics[width=\textwidth]{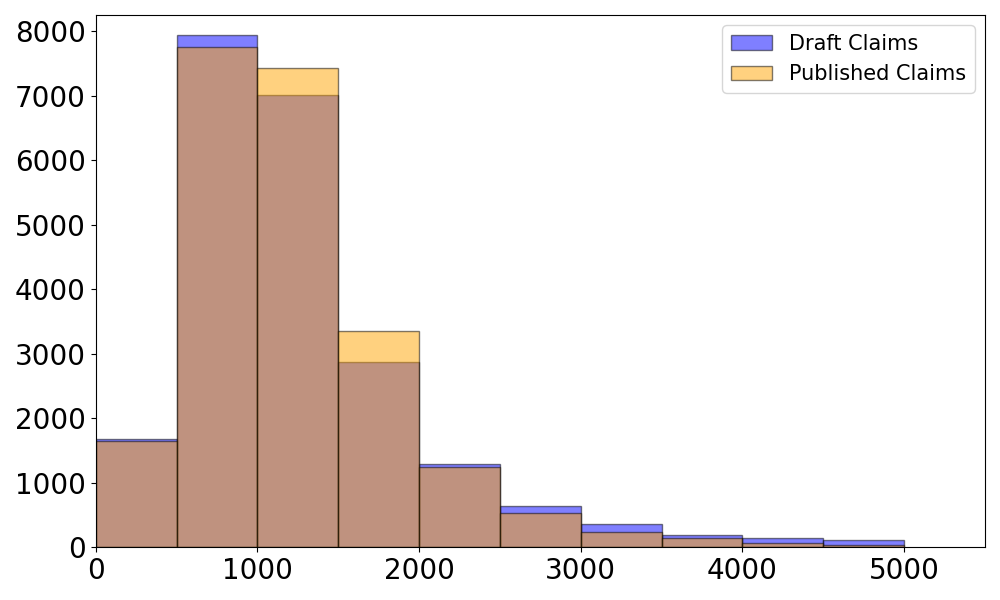}
        \caption{Number of tokens of draft claims and published claims}
        \label{fig:2}
    \end{subfigure}
    
    \caption{Frequency diagram of number of claims and tokens}
    \label{fig:stats}
\end{figure*}

\section{Dataset Statistics}

\subsection{Calculation Method}
\label{statscalculation}

We use the \textit{tiktoken} Python library and the tokenizer of GPT-3.5 to count the number of tokens. 
The claim length is calculated by the number of tokens divided by the number of claims. 
Structural complexity is determined by the ratio of subordinate clauses to the total number of sentences. We use the \textit{spaCy} Python library to analyze the number of subordinate clauses in the text. It identifies subordinate clauses by detecting dependency tags, such as csubj (clausal subject), csubjpass (clausal passive subject), ccomp (clausal complement), and xcomp (open clausal complement). Subordinate clauses increase the depth and complexity of sentence structure by adding additional information, qualifiers, or conditions. This syntactic complexity is particularly common in independent patent claims, which often incorporate numerous subordinate clauses to ensure precision and unambiguity. 
We use the Flesch-Kincaid Grade Level formula to assess the readability of the texts \citep{kincaid1975derivation}, consistent with previous studies \citep{du2022understanding}, where a lower score indicates easier readability. We use the \textit{textstat} Python library for calculation. 
The number of word changes is calculated based on \textit{difflib} Python library.

\subsection{More Statistics}
\label{morestats}
Figure~\ref{fig:stats} shows the frequency diagrams detailing the count of claims and tokens for both draft and published texts. Figures~\ref{fig:1} indicate that the number of patent claims per document predominantly ranges between 5 and 20. Figures~\ref{fig:2} reveal that claims can contain approximately 5,000 tokens, surpassing the context length limitations of some language models.

\section{Experimental Details}
\label{experimentsdetail}

All fine-tuning and inference processes are conducted on NVIDIA A100 GPUs. The total running time is about 700 hours. The following hyper-parameters are used during training: LoRA rank: 8, LoRA alpha: 16, learning rate: 5e-5, batch size: 4, number of epochs: 4, validation ratio: 10\%. 
For inference, we set the temperature to 0.1 and the maximum generation tokens to 2,048. We have employed a standard prompt format to maintain consistency. Unless otherwise specified, the input consists of prompt instruction, example claims, and a draft claim that needs revision. The following prompt instruction is used: \textit{You are a patent expert. Given the following original patent claim texts, revise claims to better withstand legal scrutiny.} We use one-shot prompting for inference, and the anticipated model output is the revised version of input claims.

\section{Model Details}
\label{modeldetail}

\textbf{CoEdIT-XL } We first evaluate the state-of-the-art text revision model, CoEdIT~\citep{raheja2023coedit}. CoEdIT models are fine-tuned Flan-T5 models \citep{chung2022scaling} based on specific data for text editing, which can output revised texts based on original texts and editing instructions. Among its variations, we opt for CoEdIT-XL due to its similar effectiveness to the larger CoEdIT-XXL model, yet with significantly fewer parameters (3 billion for XL vs. 11 billion for XXL). The maximum context length for CoEdIT is 256, but the number of tokens for most patent claims is far beyond this limit. To solve this limited token number, we segment each patent's claim set into individual claims for independent processing. Individual claims still exceeding 256 tokens are left unmodified. We use the model in a zero-shot fashion, where no training data appears in the prompt. 

\noindent \textbf{Llama-3.1 }  For open-source LLMs, we select the recent Llama-3.1, which outperforms most open-source models on common industry benchmarks \cite{dubey2024llama}. We evaluate both the Llama-3-8B-Instruct and Llama-3-70B-Instruct versions to explore the size effect. 

\noindent \textbf{Mixtral }  We also include Mixtral-8$\times$7B that is based on Sparse Mixture of Experts (SMoE) and uses the same architecture as Mistral-7B \citep{jiang2024mixtral}.  We select Mixtral-8$\times$7B-Instruct for less computational costs. 

\noindent \textbf{SaulLM } As the patent is a form of legal document, legal-specific LLMs may be useful in this task. Hence, we evaluate SaulLM-7B, a model designed for the legal domain and based on the Mistral-7B architecture \citep{colombo2024saullm}. It is trained on an English legal corpus of over 30 billion tokens, where 4.7 billion tokens are patent texts from the United States Patent and Trademark Office (USPTO). We use the SaulLM-7B-Instruct version for experiments. 

\noindent \textbf{Llama-3.1-8B-FT and SaulLM-7B-FT} We fine-tune the original Llama-3.1-8B-Instruct and SaulLM-7B model based on our train set using LoRA \citep{hu2021lora}, a parameter-efficient approach to reduce computational needs while maintaining comparable performance. The inputs are instruction prompts and the original claims. The output is revised patent claims. Appendix~\ref{experimentsdetail} lists experimental details.

\noindent \textbf{GPT-3.5 } We also include GPT series, GPT-3.5, for comparison. Specifically, we use the latest GPT-3.5 Turbo model\footnote{gpt-3.5-turbo-0125: \url{https://platform.openai.com/docs/models/gpt-3-5-turbo}}, which achieves higher accuracy in adhering to specified output formats. This model extends the context window to 16,385 tokens, which supports more examples in the prompt. Similar to the above, we evaluate GPT-3.5 with one-shot prompting, where one example is randomly chosen for each test input.  

\noindent \textbf{GPT-4 } GPT-4 represents the state-of-the-art LLM with expansive general knowledge and enhanced reasoning capabilities optimized for chat. This capability allows GPT-4 to tackle more complex challenges with increased accuracy \citep{achiam2023gpt}. We use the recent GPT-4 model\footnote{gpt-4-0125-preview: \url{https://platform.openai.com/docs/models/gpt-4-and-gpt-4-turbo}}, designed to address the problem that the model sometimes does not complete a task. GPT-4 significantly expands the context length to 128,000 tokens, so we use the same experimental setting as GPT-3.5 for fair comparison.

\begin{table*}[t!]
\footnotesize
\centering
\begin{tabular}{l|c|c|c|c|c}
\toprule
\textbf{Model} & \textbf{Completeness} & \textbf{Clarity} & \textbf{Consistency} & \textbf{Linkage} & \textbf{Quality} \\
\midrule
Copy & 81.7 & \textbf{80.8} & 79.2 & 80.5 & 80.7  \\
CoEdIT-XL  & 76.7 & 77.5 & 75.8 & 77.0 & 76.8 \\
SaulLM-7B  & \textbf{82.5} & \textbf{80.8} & 80.8 & \textbf{82.0} & \textbf{81.8}  \\
SaulLM-7B-FT  & 81.7 & 79.2 & 79.2 & 81.3 & 80.7  \\
Mixtral-8$\times$7B & \textbf{82.5} & 80.0 & \textbf{81.7} & 81.7 & 81.7  \\
Llama-3.1-8B & 79.2 & 78.3 & 80.0 & 80.0 & 79.4  \\
Llama-3.1-8B-FT & 81.7  & 79.2 & 78.3 & 80.5 & 80.3  \\
Llama-3.1-70B & 77.5 & 76.7 & 78.3 & 79.7 & 78.1 \\
GPT-3.5    & 77.5 & 76.7 & 75.0 & 77.7 & 76.9  \\
\bottomrule
\end{tabular}
\caption{GPT-4-based G-Eval evaluation results. The best score of each metric is marked in \textbf{bold}. }
\label{table:g_evaluation}
\end{table*}

\section{Evaluation Details}
\label{evaldetail}

\subsection{Human Evaluation}
\label{humanrating}

A licensed patent attorney and an experienced patent engineer, both with extensive expertise in drafting patent applications, conducted the evaluation and reached a consensus on the results. These patent professionals were provided with the referenced claims as well as those generated by LLMs. They assessed each automatically generated claim based on the criteria: Completeness of Essential Features (scored 1–10), Conceptual Clarity (scored 1–10), Consistency in Terminology (scored 1–10), and Technical Accuracy of Feature Linkages (scored 1–10). It was communicated to the evaluators that the average of their ratings would be used as the human evaluation results in the study. An ethics review board was not involved in this process.

\subsection{Standard Automated Metrics for Text Revision}
\label{standardeval}
\textbf{SARI} (System output Against References and against the Input sentence) was originally designed for text simplification tasks but is also frequently used in text revision tasks \citep{xu2016optimizing}. SARI evaluates a model's output by comparing it to both the target and the original texts, aiming to precisely assess the effectiveness in content preservation, word deletion, and word addition in the output. SARI scores range from 0 to 100. A higher score indicates better model performance.

\noindent \textbf{BLEU} (Bilingual Evaluation Understudy) quantifies the similarity between the model-generated text and the reference text through n-gram comparison \citep{papineni2002bleu}. The BLEU score, which ranges from 0 to 1, reflects the degree of correspondence between the candidate and reference texts, with scores approaching 1 indicating a higher similarity. 

\noindent \textbf{ROUGE-L} (Recall-Oriented Understudy for Gisting Evaluation -- Longest Common Sub-sequence) is designed to evaluate the generated text by measuring the longest common sub-sequence shared with the reference text, with a particular focus on the recall of the sequence \citep{lin2004rouge}. This approach aims to gauge the extent to which the model captures the essential content and maintains the structural integrity of the reference text. ROUGE-L ranges from 0 to 1, with higher values suggesting that the model has effectively preserved core contents and structure of reference materials.

\noindent \textbf{BERTScore} leverages the contextual embeddings from pre-trained transformers, such as BERT \citep{devlin2019bert}, to measure semantic similarity between the generated text and reference texts \citep{zhang2019bertscore}. BERTScore ranges from 0 to 1 and indicates the level of semantic similarity, with higher values denoting greater similarity

For automated evaluation metrics, we use the package from the HuggingFace \textit{evaluate} library.\footnote{\url{https://github.com/huggingface/evaluate}} 

\subsection{G-Eval}
\label{gevaldetails}

We use the following prompt for G-Eval. \textit{You will be given the draft claims and the referenced claims of the same patent. \
Your task is to rate the draft claims on four metrics using the referenced claims as the gold standard. \
Please make sure you read and understand these instructions carefully. Please keep this document open while reviewing, and refer to it as needed. \
Evaluation Criteria: \
1. Completeness of Essential Features (0-100): The extent to which the generated claims encapsulated all critical aspects of the invention. \
- 0-20: Most essential features are missing or poorly described. \
- 21-40: Some essential features are present but significant gaps remain. \
- 41-60: Majority of essential features are covered but with minor omissions. \
- 61-80: Almost all essential features are well described with very few gaps. \
- 81-100: All essential features are thoroughly and comprehensively covered. \
2. Conceptual Clarity (0-100): The clarity and unambiguity of the language used in the claims. \
- 0-20: Claims are very unclear and ambiguous.\
- 21-40: Claims have significant clarity issues, making them difficult to understand.\
- 41-60: Claims are mostly clear but contain some ambiguous language.\
- 61-80: Claims are clear with minimal ambiguity.\
- 81-100: Claims are exceptionally clear and completely unambiguous. \
3. Consistency in Terminology (0-100): The uniformity in the use of terms throughout the claims. \
- 0-20: Terminology is highly inconsistent.\
- 21-40: Significant inconsistencies in terminology.\
- 41-60: Some inconsistencies in terminology but mostly uniform.\
- 61-80: Terminology is largely consistent with minor inconsistencies.\
- 81-100: Terminology is completely consistent throughout.\
4. Technical Correctness of Feature Linkages (0-100): The accuracy with which the features were interconnected and related.\
- 0-20: Features are poorly linked with many inaccuracies.\
- 21-40: Significant issues with the linkages of features.\
- 41-60: Mostly accurate linkages with some incorrect connections.\
- 61-80: Accurate linkages with minor inaccuracies.\
- 81-100: Features are accurately and correctly linked throughout.\
Evaluation Steps: \
1. Read the referenced claims carefully and identify the inventions' features. Assume the referenced claims have scores of 100 in all Evaluation Criteria. \
2. Read the draft claims and compare it to the referenced claims. \
3. Assign a score for each metric based on the Evaluation Criteria. \
Example: \
Referenced Claims:  \
<<Claims>> \
Draft Claims:  \
<<Claims>> \
Evaluation Form (scores ONLY): \
- Completeness of Essential Features: X, - Conceptual Clarity: X, - Consistency in Terminology: X, - Technical Correctness of Feature Linkages: X.  }

We use GPT-4 to obtain the scores of completeness of essential features, conceptual clarity, consistency in terminology, and technical correctness of feature linkages. The overall quality is calculated based on the same formula of human evaluation.

\section{More Results}
\label{moreresult}

In this study, we also employ the original version of Llama-2, which has not been fine-tuned for chat-based interactions or question-answering tasks. Therefore, the revised claim is the natural continuation of the input prompt, leading to some potential issues. Following the revised claims generated by Llama-2, we observe instances where the output continues to include claims not found in either the training or testing datasets, likely a result of its inclusion during the pre-training phase. In line with findings from \citet{raheja2023coedit}, we note that Llama-2 tends to replicate the input without modification. Furthermore, we find that some of the output claims were incomplete, abruptly ending mid-generation. These findings suggest that Llama-2 without instruction-tuning may struggle with accurately interpreting the prompted task, leading to repetitive or irrelevant outputs.

Table~\ref{stat} shows that GPT-3.5 notably reduced the average token count from 1,124 to 831. Compared to other models, GPT-3.5 generates the shortest claim length of 60 and exhibits the lowest structure complexity of 0.67. Therefore, the result demonstrates that GPT-3.5 prefers straightforward language and simple sentence structures when revising claims, a strategy that fails to meet the stringent requirements of patent claims. In human evaluation, the claim quality score of 5.6 from GPT-3.5 does not surpass the copy baseline, indicating that the edits are not markedly effective. 

We report the full results of G-Eval in Table \ref{table:g_evaluation} for references. 

\section{Broader Impact}
The new task of patent claim revision and related challenges provide valuable insights for the broader field of text revision. Current text-revision models cannot perform well on patent claim revision, because of the unique characteristics of patent texts and goals of revision. This task offers a compelling case for further research into different types and domain-specific text revision methodologies. In addition, the results demonstrate the importance of fine-tuning and domain adaptation for this patent task. Compared to specified LLMs, general-purpose LLMs show limited effectiveness, a finding that may apply to other related fields such as legal documentation. Furthermore, the results underscore the limitations of current automated evaluation metrics in structured, domain-specific contexts. The gap between human assessments and automated metrics highlights the need for more robust evaluation methods that better align with human judgment in specialized contexts.

\end{document}